%% file: DepthPruning.tex
\documentclass[dvipsnames]{article}
\usepackage{spconf,amsmath,graphicx}
\usepackage{array}
\usepackage{pgfplots}

\usepackage{fancyhdr}

\usepackage{algorithm}
\usepackage{algpseudocode}
\usepackage[implicit=false]{hyperref}
\hypersetup{
    colorlinks=true,
    linkcolor=blue,
    filecolor=magenta,      
    urlcolor=black,
    }

\pgfplotsset{compat=1.3}
\pgfplotsset{
  tick label style = {font=\sansmath\sffamily},
  every axis label = {font=\sansmath\sffamily},
  legend style = {font=\sansmath\sffamily},
  label style = {font=\sansmath\sffamily},
}

\usetikzlibrary{pgfplots.groupplots}
\usetikzlibrary{decorations.text}
\usetikzlibrary{arrows.meta}

\usepackage[eulergreek]{sansmath}

\pgfplotsset{tikzDefaults/.style=
  {
      font=\sffamily\sansmath
  }
}

\title{Depth Pruning with Auxiliary Networks for TinyML}
\name{Josen Daniel De Leon$^{\star \dagger}$, Rowel Atienza$^{\star}$}
\address{
        $^{\star}$Electrical and Electronics Engineering Institute, University of the Philippines \\
        $^{\dagger}$Samsung Research Philippines \\
        \{josen.daniel.de.leon, rowel\}@eee.upd.edu.ph}

\fancypagestyle{FirstPage}{
\lfoot{\small\copyright 2022 IEEE. Personal use of this material is permitted. Permission from IEEE must be obtained for all other uses, in any current or future media, including reprinting/republishing this material for advertising or promotional purposes, creating new collective works, for resale or redistribution to servers or lists, or reuse of any copyrighted component of this work in other works.} 
}

\begin{document}
%
\maketitle
\begin{abstract}

Pruning is a neural network optimization technique that sacrifices accuracy in exchange for lower computational requirements. Pruning has been useful when working with extremely constrained environments in tinyML. Unfortunately, special hardware requirements and limited study on its effectiveness on already compact models prevent its wider adoption. Depth pruning is a form of pruning that requires no specialized hardware but suffers from a large accuracy falloff. To improve this, we propose a modification that utilizes a highly efficient auxiliary network as an effective interpreter of intermediate feature maps. Our results show a parameter reduction of $93\%$ on the MLPerfTiny Visual Wakewords (VWW) task and $28\%$ on the Keyword Spotting (KWS) task with accuracy cost of $0.65\%$ and $1.06\%$ respectively. When evaluated on a Cortex-M0 microcontroller, our proposed method reduces the VWW model size by $4.7\times$ and latency by $1.6\times$ while counter intuitively gaining $1\%$ accuracy. KWS model size on Cortex-M0 was also reduced by $1.2\times$ and latency by $1.2\times$ at the cost of $2.21\%$ accuracy.


\end{abstract}

\section{Introduction}
\label{sec:intro}
\input{sections/1-introduction.tex}
\thispagestyle{FirstPage}

\section{Related Work}
\label{sec:relatedwork}
\input{sections/2-relatedworks.tex}

\section{Depth Pruning with Auxiliary Networks}
\label{sec:methodology}
\input{sections/3-methodology.tex}

\section{Experiments and Results}
\label{sec:experiments}
\input{sections/4-results.tex}

\section{Conclusions and Future Work}
\label{sec:conclusion}
\input{sections/5-conclusion.tex}

\vfill\pagebreak



\bibliographystyle{IEEEbib}
\bibliography{strings,refs}

\end{document}

%% file: sections/1-introduction.tex
There has been a recent growing interest to bring machine learning inference to commercial devices driven by the increasing consumer interest in privacy, energy efficiency and autonomy of edge devices. Market studies speculate that shipments of smart devices making use of these developments could grow from 15.2M in 2020 to 2.5B in 2030 \cite{ABIResearch2021}.

New challenges have emerged from this movement towards more ubiquitous devices. These mainly stem from the extreme constraints on compute resources imposed by ultra-low power devices, typically under the mW range. As such, academe and industry leaders have established tinyML\cite{tinyMLWebsite2021} as a field focused on the optimization of the different stages of on-device inference in extremely constrained environments. These may impose model sizes in the kilobytes (KB) range and operate with only MFLOPS of compute power.

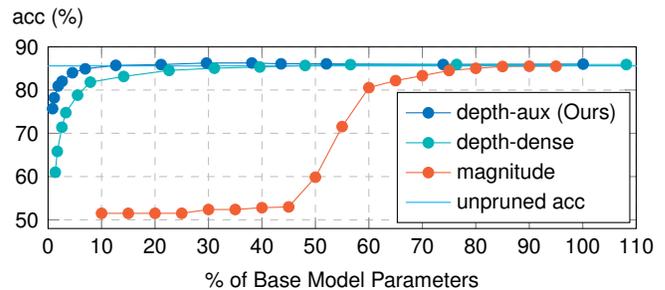
\begin{figure}
    \centering

    \begin{tikzpicture}[] 
        \begin{axis}[
            height = 4cm,
            width = 9.4cm,
            xlabel={\% of Base Model Parameters},
            ymin=48, ymax=90,
            ytick scale label code/.code={},
            xtick scale label code/.code={},
            xtick={0,10,20,30,40,50,60,70,80,90,100,110},
            xmin=0,xmax=110,
        	ylabel=acc (\%),
            ymajorgrids=true,
            xmajorgrids=true,
            grid style=dashed,
            ticklabel style = {font=\footnotesize\sffamily\sansmath},
            label style = {font=\footnotesize\sffamily\sansmath},
            y label style = {at={(axis description cs:0,1.05)},rotate=-90,anchor=south},
            legend style = {font=\footnotesize\sffamily\sansmath,at={(0.99,0.03)},anchor=south east},
            legend cell align={left}
        ]
        \addplot[mark=*, color=RoyalBlue]
            table[x=percent_params,y=acc]{data/vww_mobilenet_aux-pruning.dat};
        \addlegendentry{depth-aux (Ours)}
        \addplot[mark=*, color=BlueGreen]
            table[x=percent_params,y=acc]{data/vww_mobilenet_dense-pruning.dat};
        \addlegendentry{depth-dense}
        \addplot[mark=*, color=RedOrange]
            table[x=percent_params,y=acc]{data/vww_mobilenet_mag-pruning.dat};
        \addlegendentry{magnitude}
        \addplot[color=cyan]
        coordinates {
        (-10,85.57)(120,85.57)
        };
        \addlegendentry{unpruned acc}
        \end{axis}
    \end{tikzpicture} 

    
    \caption{Pruning Method Comparison on MobilenetV1 VWW task. Parameters are reduced by $93\%$ for our method (frozen tail), $68.9\%$ for depth pruning (dense), and $20\%$ for magnitude pruning if maximum accuracy drop is set to $0.65\%$.}
    
    \label{figure:depthpruning_comp}
\end{figure}

We contribute to this space by exploring how effective depth pruning is on tinyML tasks. Depth pruning is a technique where entire layers are removed from the end of a trained model until a target model size is reached. We also introduce a modification to this method by using a small auxiliary network as a new head to improve accuracy with minimal overhead. This method is well structured and has no special hardware requirements. Our experiments show a parameter reduction of $93\%$ with accuracy cost of $0.65\%$ on the MLPerfTiny Visual Wakewords (VWW) task as shown in Figure \ref{figure:depthpruning_comp}. We validate our results by deploying the pruned Visual Wakewords (VWW) model on an inexpensive (\$$3$ retail \cite{Mouser2021}) and widely used ARM Cortex-M0. This results to a $1.6\times$ on-device inference speedup and a $4.7\times$ smaller model size while counter intuitively gaining $1\%$ in accuracy.

%% file: sections/2-relatedworks.tex
\subsection{TinyML, Use Cases and Hardware}
TinyML is an emerging field in machine learning with the goal of bringing machine learning to edge devices. Various benchmarks \cite{Banbury2021, Banbury2020} have been selected to represent tinyML applications and use cases. Among them is the Visual Wakewords (VWW) task which is a person presence detection problem using $115$k train and $8$k validation $96\times96$ RGB images \cite{Chowdhery2019}. Another is the Keyword Spotting (KWS) task which uses a dataset with $105,829$ 1sec word utterances such as "up", "down", "yes" and "no" to define a speech recognition problem with $12$ classes \cite{Warden2018,Zhang2017}. 


Microcontrollers provide an attractive and low cost hardware environment to achieve the goals of tinyML \cite{Zhang2017, Banbury2020-2}. Unfortunately, they also have significantly low compute capabilities as seen in Table \ref{table:microcontrollers}. This influences neural network design for microcontrollers to become a multi-objective optimization problem \cite{Fedorov2019} between performance and computate requirements. As such, methods which allow us to tweak models towards Pareto-optimal allocations between these metrics become important to tinyML research.

\begin{table}
    \centering
    \small
    \begin{tabular}{m{10em} m{5em} >{\raggedleft\arraybackslash}m{3em} >{\raggedleft\arraybackslash}m{3em}} 
     \hline
     \centering\arraybackslash Platform & \centering\arraybackslash Processor & \centering\arraybackslash FLOPS & \centering\arraybackslash RAM \\
     \hline
      \hline
     Desktop & RTX 3080Ti & 34 T & 128GB \\
      \hline
     Galaxy Note 20 (Mobile) & Exynos 990 & 1.1 T & 8GB \\
       \hline
       \hline
     RPi 4 Model B & Cortex-A72 & 48 G & 4GB \\
       \hline
     Arducam Pico4ML & Cortex-M0 & 16 M & 264KB \\
     \hline
    \end{tabular}
    \caption[Comparison of compute resources of Desktop, Mobile and Cortex Microcontrollers]{Comparison of compute resources of Desktop, Mobile and Cortex Microcontrollers}
    \label{table:microcontrollers}
\end{table}

\subsection{Network Pruning} 
One avenue of optimization in tinyML involves the modification of trained model parameters to reduce or simplify the computations during inference. Pruning \cite{LeCun1990,Han2015,Blalock2020} is one such method wherein unimportant parameters from a trained model are removed to reduce model size. It relies on a hypothesis that there exists a "winning lottery ticket" subnetwork \cite{Frankle2019, Alabdulmohsin2021} that allows us to reach the same accuracy as the original network using only a subset of its parameters. Despite progress in this research, identifying an effective pruning scheme for a specific model can still be considered as an unstructured process requiring iteration and intuition.

Among pruning schemes that aim to identify this subnetwork, magnitude pruning\cite{Blalock2020, Zhu2017} can be seen as the most popular . This method involves using a parameter's magnitude as a heuristic to determine which weights can be zeroed out to produce a sparse model. Sparse matrix computation and compression then lowers compute requirements during inference. Unfortunately, this feature is rarely found in microcontrollers limiting the adoption of pruning in commercial applications.

\subsection{Depth Pruning}

Depth pruning is a form of pruning where layers at the end of a trained model are removed to reduce model size. We define depth pruning as a technique that generates a subnetwork from a trained base model with $N$ layers. We retain the first $D$ layers from the base model and attach a new classifier head, typically a single dense layer, after $D$. 

A prior work in intermediate feature map interpretability \cite{Alain2016} shows that this method is able to produce an accuracy curve that is monotonically increasing with model depth. This property, along with most neural networks designs having parameters concentrated at deeper layers, gives us an easy-to-use and interpretable knob to adjust the trade-off between accuracy and compute requirements.

While this approach shares similarities with transfer learning, differences lie in their objectives. Whereas transfer learning uses stored knowledge on a certain task to improve performance on a different downstream task, depth pruning uses it to produce smaller models on the same task. In addition, transfer learning usually removes only the final layer while depth pruning aims to discard as many as possible.

While depth pruning has been shown to function well with Neural Architecture Search \cite{Cai2019} and requires no special hardware, it is seldom used as a standalone solution due to the large falloff in accuracy. We hypothesize that this can be improved by using a highly efficient auxiliary network instead of just a single dense layer.

%% file: sections/3-methodology.tex
\begin{figure}[h]
    \centering
    \includegraphics[width=0.35\textwidth]{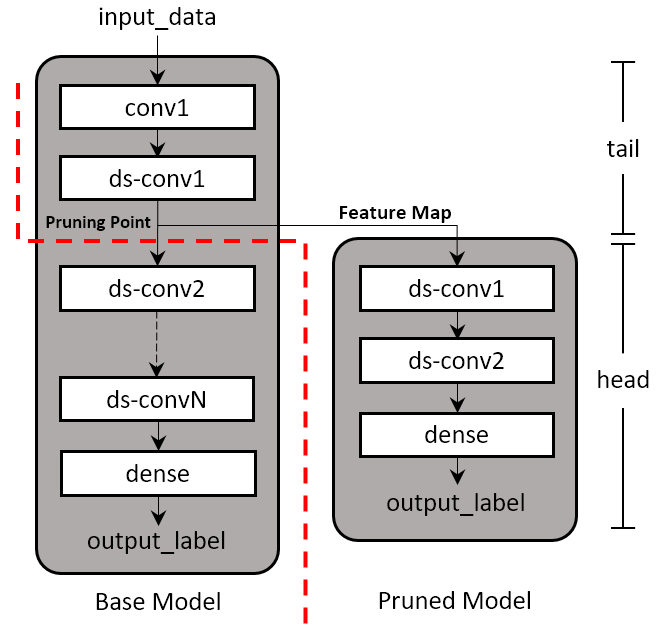}
    \caption[Depth Pruning with Auxiliary Networks]{Depth Pruning with Auxiliary Networks. Our pruning method applied to an arbitrary trained model.}
    \label{fig:arch}
\end{figure}

Our work improves the accuracy of models produced by depth pruning through the use of an auxiliary network as the new layer head as illustrated in Figure \ref{fig:arch}. This acts as a powerful interpreter of intermediate feature maps from the trained layers to the output labels. Furthermore, this auxiliary network incurs minimal overhead and is simple to train and apply. Model training and pruning code is located at \href{https://github.com/jd-deleon/depth-pruning-auxnets}{https://github.com/jd-deleon/depth-pruning-auxnets}

Table \ref{table:aux_arch} outlines the architecture of the auxiliary network selected for this study. We make use of depthwise separable convolutions which have been shown to be highly efficient building blocks for compact neural networks \cite{Howard2017}. Hyperparameters are selected such that the auxiliary network incurs minimal overhead while still being able to preserve accuracy. We make use of ReLU activation and batch normalization after each convolution to stabilize training.

\begin{table}
    \centering
    \small
    \begin{tabular}{m{6em} >{\centering\arraybackslash}m{6em} >{\raggedleft\arraybackslash}m{6em}} 
     \hline
     \centering\arraybackslash Type / Stride & \centering\arraybackslash Filter Shape & \centering\arraybackslash Parameters \\
     \hline
      \hline
     dw-Conv / s1 & 3$\times$3$\times$[64] dw & [896] \\
      \hline
     pw-Conv / s1 &  1$\times$1$\times$[64]$\times$32 & [2,208] \\
       \hline
     dw-Conv / s1 & 3$\times$3$\times$32 dw & [448] \\ 
     \hline
     pw-Conv / s1 & 1$\times$1$\times$32$\times$16 & [592]  \\
     \hline
     globalAvePool & - & 0  \\
     \hline
     dense & 32$\times$[2] & [34]  \\
     \hline
    \end{tabular}
    \caption[Auxiliary Network Architecture]{Auxiliary Network Architecture. Bracketed values are dependent on the input shape. Presented values are with a 12$\times$12$\times$64 tensor from MobilenetV1-VWW block5.}
    \label{table:aux_arch}
\end{table}

Algorithm \ref{algorithm:depth_pruning_ax} outlines how our pruning method is conducted in a trained model. In summary, we take the first $p$ layers of our trained model $W$ as our tail and freeze it as an optional step. We then attach our auxiliary network as the new head and train it to produce the pruned model.

\begin{algorithm}
\small
\caption{Depth Pruning with Auxiliary Networks}\label{alg:cap}
\begin{algorithmic}[1]
\Require $W$, the trained model layers, $p$, the pruning point, and
$X$, the dataset on which to finetune
\State $T \gets W[0:p]$
\State $T \gets freeze(T)$
\State $A \gets initializeAux()$
\State $W' \gets append(T,A)$
\State $W' \gets trainToConvergence(f(X;W'))$ \\
\Return $W'$
\end{algorithmic}
\label{algorithm:depth_pruning_ax}
\end{algorithm}

Freezing the model tail is an optional step which produces a pruned model having shared weights with the base network. This can be used in dynamic architectures wherein a smaller network can signal the activation of a larger high performance branch. We can also let the tail remain unfrozen allowing the model to converge to a better accuracy. 

We retain all the data preprocessing, optimizer settings, loss functions and labelling conventions used to train the base network during the pruning process. This takes advantage of any prior knowledge already learned by the tail of the model. While training the auxiliary network usually takes only a few epochs, we train for the same number of epochs as in base model training to ensure convergence in the reported data.

%% file: sections/4-results.tex
We use TensorFlow2 \cite{tensorflow2021} model definitions and code samples in the MLPerfTiny benchmarking suite \cite{Banbury2021} to train and evaluate the base networks for both VWW and KWS tasks. We use step learning rates $[0.001,0.0005,0.00025]$ for $[20,10,20]$ epochs on VWW and $[0.0005,0.0001,0.00002]$ for $[12,12,6]$ epochs on KWS in the Adam \cite{Kingma2015} optimizer. All figures include overheads from the auxiliary network which may result to some configurations exceeding base model properties.

\subsection{Depth Pruning on tinyML Tasks}
\begin{figure}
    \centering
    \begin{tikzpicture}[]
    
    \begin{groupplot}[group style = {group size = 2 by 1, horizontal sep = 29pt}, width = 4.9cm, height = 4cm]
    
    \nextgroupplot[
        xlabel={Pruning Location},
        xticklabels from table={data/vww_aux-pruning.dat}{attachment_point},
        x tick label style={rotate=60,anchor=east},
        axis y line*=right,
        ymin=0, ymax=300000,
        xtick=data,
        ytick scale label code/.code={},
        ytick={0, 50000, 100000, 150000, 200000, 250000, 300000},
        yticklabels={0, 50, 100, 150, 200, 250, 300},
        xmin=-0.5,xmax=13.5,
    	ylabel=Params (K),
        ybar,
        bar width=6.7,
        ymajorgrids=true,
        xmajorgrids=true,
        grid style=dashed,
        ticklabel style = {font=\scriptsize\sffamily\sansmath},
        label style = {font=\scriptsize\sffamily\sansmath},
        y label style = {at={(axis description cs:0.95,1.0)},rotate=-90,anchor=south}
    ]
    
    \addplot[fill=Gray]
        table[x expr=\coordindex,y=parameters]{data/vww_aux-pruning.dat};

    \nextgroupplot[
        xlabel={Pruning Location},
        xticklabels from table={data/vww_aux-pruning.dat}{attachment_point},
        x tick label style={rotate=60,anchor=east},
        axis y line*=right,
        ymin=0, ymax=30000000,
        xtick=data,
        ytick scale label code/.code={},
        ytick={0, 5000000, 10000000, 15000000, 20000000,25000000,30000000},
        yticklabels={0, 5, 10, 15, 20, 25,30},
        xmin=-0.5,xmax=13.5,
    	ylabel=FLOPS (M),
        ybar,
        bar width=6.7,
        ymajorgrids=true,
        xmajorgrids=true,
        grid style=dashed,
        ticklabel style = {font=\scriptsize\sffamily\sansmath},
        label style = {font=\scriptsize\sffamily\sansmath},
        y label style = {at={(axis description cs:0.95,1.0)},rotate=-90,anchor=south}
    ]
    
    \addplot[fill=Gray]
        table[x expr=\coordindex,y=flops]{data/vww_aux-pruning.dat};
    \end{groupplot}
    
    \begin{groupplot}[group style = {group size = 2 by 1, horizontal sep = 29pt}, width = 4.9cm, height = 4cm]
    
    \nextgroupplot[
        axis y line*=left,
        axis x line=none,
        ylabel={acc (\%)},
        ymin=70, ymax=90,
        xtick=data,
        ytick={70,75,80,85,90},
        xmin=-0.5,xmax=13.5,
        ticklabel style = {font=\scriptsize\sffamily\sansmath,color=RoyalBlue},
        label style = {font=\scriptsize\sffamily\sansmath,color=RoyalBlue},
        legend style = { column sep = 0pt, legend columns = -1, legend to name = grouplegend_vww,font=\scriptsize\sffamily\sansmath},
        legend cell align={left},
        axis line style={RoyalBlue},
        y label style = {at={(axis description cs:0,1.0)},rotate=-90,anchor=south}
        ]
        
    ]
    \addplot[mark=*, color=RoyalBlue, mark size=1.5pt]
        table[
        x expr=\coordindex,y=accuracy
        ] 
        {data/vww_aux-pruning.dat};
    \addlegendentry{frozen tail}
    \addplot[mark=*, color=BlueGreen, mark size=1.5pt]
        table[
        x expr=\coordindex,y=accuracy_unfrozen
        ] 
        {data/vww_aux-pruning.dat};
    \addlegendentry{unfrozen tail}
    \addplot[color=cyan]
        coordinates {
        (-1,85.57)(14,85.57)
        };
    
    \addlegendentry{unpruned acc}
    
 \draw[red, thick] (axis cs:5, \pgfkeysvalueof{/pgfplots/ymin}) -- (axis cs:5, \pgfkeysvalueof{/pgfplots/ymax});
    \draw[red, thick] (axis cs:13, \pgfkeysvalueof{/pgfplots/ymin}) -- (axis cs:13, \pgfkeysvalueof{/pgfplots/ymax});
    \draw[red, very thick, {Stealth}-{}] (axis cs:5,84.9) -- (axis cs:13,85.57);
    \node[align=left,font=\scriptsize\sffamily\sansmath\bfseries] at (axis cs:8.5,85.3) [anchor=north] {\textcolor{RoyalBlue}{acc}$\downarrow0.65\%$\\ param:$\downarrow93\%$};
    
    \nextgroupplot[
        axis y line*=left,
        axis x line=none,
        ylabel={acc (\%)},
        ymin=70, ymax=90,
        xtick=data,
        ytick={70,75,80,85,90},
        xmin=-0.5,xmax=13.5,
        ticklabel style = {font=\scriptsize\sffamily\sansmath,color=RoyalBlue},
        label style = {font=\scriptsize\sffamily\sansmath,color=RoyalBlue},
        axis line style={RoyalBlue},
        y label style = {at={(axis description cs:0,1.0)},rotate=-90,anchor=south}
    ]
    
     \draw[red, thick] (axis cs:5, \pgfkeysvalueof{/pgfplots/ymin}) -- (axis cs:5, \pgfkeysvalueof{/pgfplots/ymax});
    \draw[red, thick] (axis cs:13, \pgfkeysvalueof{/pgfplots/ymin}) -- (axis cs:13, \pgfkeysvalueof{/pgfplots/ymax});
    \draw[red, very thick, {Stealth}-{}] (axis cs:5,84.9) -- (axis cs:13,85.57);
    \node[align=left,font=\scriptsize\sffamily\sansmath\bfseries] at (axis cs:8.6,85.3) [anchor=north] {\textcolor{RoyalBlue}{acc}$\downarrow0.65\%$\\ FLOPS:$\downarrow51\%$};
    
    \addplot[mark=*, color=RoyalBlue, mark size=1.5pt]
        table[
        x expr=\coordindex,y=accuracy
        ] 
        {data/vww_aux-pruning.dat};
        
    \addplot[mark=*, color=BlueGreen, mark size=1.5pt]
        table[
        x expr=\coordindex,y=accuracy_unfrozen
        ] 
        {data/vww_aux-pruning.dat};
    \addplot[color=cyan]
        coordinates {
        (-1,85.57)(14,85.57)
        };
        
    \end{groupplot}
    \node at ($(group c2r1) + (-2.2cm,-3.2cm)$) {\ref{grouplegend_vww}}; 
    \end{tikzpicture} 
    
    \caption{Depth Pruning on MobilenetV1 VWW task. Pruning at block5 reduces parameters by $93\%$ and FLOPS by $51\%$ at a cost of $0.65\%$ accuracy.}
    \label{figure:depthpruning_vww_param}
\end{figure}
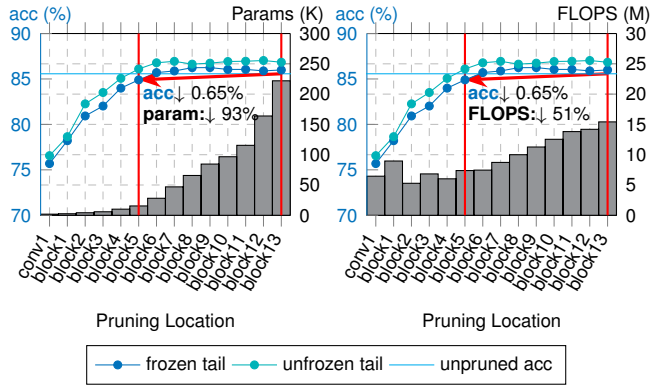

From our VWW results in Figure \ref{figure:depthpruning_vww_param}, we observe that our method can significantly reduce operational requirements with minimal cost to accuracy. We also notice that some pruning locations are able to produce smaller models with higher accuracy than the base model. This may be attributed to overparametrization in the base model leading to overfitting which is reduced by our pruning process.

\begin{figure}
    \centering

    \begin{tikzpicture}[] 
    \begin{groupplot}[group style = {group size = 2 by 1, horizontal sep = 29pt}, width = 4.9cm, height = 4cm]

        \nextgroupplot[
            xlabel={Pruning Location},
            axis y line*=right,
            xticklabels from table={data/kws_aux-pruning.dat}{attachment_point},
            ymin=0, ymax=60000,
            xtick=data,
            ytick scale label code/.code={},
            ytick={0, 10000,20000, 30000,40000, 50000, 60000},
            yticklabels={0, 10, 20, 30, 40, 50, 60},
            xmin=-0.5,xmax=4.5,
            x tick label style={rotate=60,anchor=east},
        	ylabel=Params (K),
            ybar,
            bar width=0.6cm,
            ymajorgrids=true,
            xmajorgrids=true,
            grid style=dashed,
            ticklabel style = {font=\scriptsize\sffamily\sansmath},
            label style = {font=\scriptsize\sffamily\sansmath},
            y label style = {at={(axis description cs:0.92,1.0)},rotate=-90,anchor=south}
        ]
        \addplot[fill=Gray]
            table[x expr=\coordindex,y=parameters]{data/kws_aux-pruning.dat};
    
        \nextgroupplot[
            xlabel={Pruning Location},
            axis y line*=right,
            xticklabels from table={data/kws_aux-pruning.dat}{attachment_point},
            ymin=0, ymax=12000000,
            xtick=data,
            ytick scale label code/.code={},
            ytick={0, 2000000, 4000000, 6000000, 8000000, 10000000, 12000000},
            yticklabels={0, 2, 4, 6, 8, 10, 12},
            xmin=-0.5,xmax=4.5,
            x tick label style={rotate=60,anchor=east},
        	ylabel=FLOPS (M),
            ybar,
            bar width=0.6cm,
            ymajorgrids=true,
            xmajorgrids=true,
            grid style=dashed,
            ticklabel style = {font=\scriptsize\sffamily\sansmath},
            label style = {font=\scriptsize\sffamily\sansmath},
            y label style = {at={(axis description cs:0.94,1.0)},rotate=-90,anchor=south}
        ]
        
        \addplot[fill=Gray]
            table[x expr=\coordindex,y=flops]{data/kws_aux-pruning.dat};

    \end{groupplot}
    
    \begin{groupplot}[group style = {group size = 2 by 1, horizontal sep = 29pt}, width = 4.9cm, height = 4cm]
        
        \nextgroupplot[
            xlabel={pruning\_point},
            axis x line=none,
            axis y line*=left,
            ylabel={acc (\%)},
            ymin=88, ymax=97,
            xtick=data,
            ytick={88,90,92,94,96},
            xmin=-0.5,xmax=4.5,
            x tick label style={rotate=60,anchor=east},
            ticklabel style = {font=\scriptsize\sffamily\sansmath,color=RoyalBlue},
            label style = {font=\scriptsize\sffamily\sansmath,color=RoyalBlue},
            legend style = { column sep = 0pt, legend columns = -1, legend to name = grouplegend_kws,font=\scriptsize\sffamily\sansmath},
            legend cell align={left},
            axis line style={RoyalBlue},
            y label style = {at={(axis description cs:0,1.0)},rotate=-90,anchor=south}
        ]
        
        \draw[red, thick] (axis cs:2, \pgfkeysvalueof{/pgfplots/ymin}) -- (axis cs:2, \pgfkeysvalueof{/pgfplots/ymax});
        \draw[red, thick] (axis cs:4, \pgfkeysvalueof{/pgfplots/ymin}) -- (axis cs:4, \pgfkeysvalueof{/pgfplots/ymax});
        \draw[red, very thick, {Stealth}-{}] (axis cs:2,94.65) -- (axis cs:4,95.71);
        \node[align=left,font=\scriptsize\sffamily\sansmath\bfseries] at (axis cs:3,94.5) [anchor=north] {\textcolor{RoyalBlue}{acc}$\downarrow1.06\%$\\ params:$\downarrow28\%$};
        
        \addplot[mark=*, color=RoyalBlue]
            table[
            x expr=\coordindex,y=accuracy
            ] 
            {data/kws_aux-pruning.dat};
        \addlegendentry{frozen tail}
        \addplot[mark=*, color=BlueGreen]
            table[
            x expr=\coordindex,y=accuracy_unfrozen
            ] 
            {data/kws_aux-pruning.dat};
        \addlegendentry{unfrozen tail}
        \addplot[
            color=cyan,
            ]
            coordinates {
            (-1,95.71)(5,95.71)
            };
        \addlegendentry{unpruned acc}
        

        \nextgroupplot[
            xlabel={pruning\_point},
            axis x line=none,
            axis y line*=left,
            ylabel={acc (\%)},
            ymin=88, ymax=97,
            xtick=data,
            ytick={88, 90,92,94,96},
            xmin=-0.5,xmax=4.5,
            ticklabel style = {font=\scriptsize\sffamily\sansmath,color=RoyalBlue},
            label style = {font=\scriptsize\sffamily\sansmath,color=RoyalBlue},
            axis line style={RoyalBlue},
            y label style = {at={(axis description cs:0,1.0)},rotate=-90,anchor=south}
        ]
        
        \draw[red, thick] (axis cs:2, \pgfkeysvalueof{/pgfplots/ymin}) -- (axis cs:2, \pgfkeysvalueof{/pgfplots/ymax});
        \draw[red, thick] (axis cs:4, \pgfkeysvalueof{/pgfplots/ymin}) -- (axis cs:4, \pgfkeysvalueof{/pgfplots/ymax});
        \draw[red, very thick, {Stealth}-{}] (axis cs:2,94.65) -- (axis cs:4,95.71);
        \node[align=left,font=\scriptsize\sffamily\sansmath\bfseries] at (axis cs:3.3,95) [anchor=north] {\textcolor{RoyalBlue}{acc}$\downarrow1.06\%$\\ FLOPS:$\downarrow28\%$};

        \addplot[mark=*, color=RoyalBlue]
            table[
            x expr=\coordindex,y=accuracy
            ] 
            {data/kws_aux-pruning.dat};
        \addplot[mark=*, color=BlueGreen]
            table[
            x expr=\coordindex,y=accuracy_unfrozen
            ] 
            {data/kws_aux-pruning.dat};
        \addplot[
            color=cyan,
            ]
            coordinates {
            (-1,95.71)(5,95.71)
            };

    \end{groupplot}
    \node at ($(group c2r1) + (-2.2cm,-3.1cm)$) {\ref{grouplegend_vww}}; 
    \end{tikzpicture} 

    \caption{Depth Pruning on DSCNN KWS task. Pruning at block2 reduces parameters by $28\%$ and FLOPS by $28\%$ at a cost of $1.06\%$ accuracy.}
    \label{figure:depthpruning_kws_param}
\end{figure}
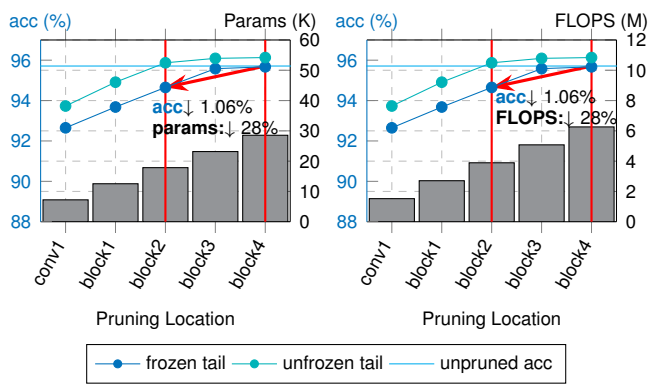

Our results with the much smaller models for the KWS task in Figure \ref{figure:depthpruning_kws_param} is able to yield less improvement due to higher information density in compact models. In spite of this, we are still able to produce smaller models with minimal loss in accuracy. Unfreezing our tail is able to produce models with higher accuracy on both tasks. Latency measurements were conducted on an Intel-i9 CPU but produced no noticeable improvement from $5$ms due to the already compact model sizes prior to pruning. 

\begin{figure}
    \centering

    \begin{tikzpicture}[] 
        \begin{axis}[
            height = 4.25cm,
            width = 9.4cm,
            xlabel={\% of Base Model Parameters},
            ymin=55, ymax=100,
            ytick scale label code/.code={},
            xtick scale label code/.code={},
            xtick={0,10,20,30,40,50,60,70,80,90,100,110,120},
            xmin=0,xmax=130,
        	ylabel=acc (\%),
            ymajorgrids=true,
            xmajorgrids=true,
            grid style=dashed,
            ticklabel style = {font=\footnotesize\sffamily\sansmath},
            label style = {font=\footnotesize\sffamily\sansmath},
            y label style = {at={(axis description cs:0,1.05)},rotate=-90,anchor=south},
            legend style = {font=\footnotesize\sffamily\sansmath,at={(0.99,0.03)},anchor=south east, legend columns = 2},
            legend cell align={left}
        ]
        \addplot[color=cyan]
        coordinates {
        (-10,95.71)(130,95.71)
        };
        \addlegendentry{unpruned acc}
        \addplot[mark=*, color=RedOrange]
            table[x=percent_params,y=acc]{data/kws_dscnn_mag-pruning.dat};
        \addlegendentry{magnitude}
        \addplot[mark=*, color=BlueGreen]
            table[x=percent_params,y=acc]{data/kws_dscnn_dense-pruning.dat};
        \addlegendentry{depth-dense}
        \addplot[mark=*, color=RoyalBlue]
            table[x=percent_params,y=acc]{data/kws_dscnn_aux-pruning.dat};
        \addlegendentry{depth-aux (Ours)}

        \end{axis}
    \end{tikzpicture} 

    
    \caption{Pruning Method Comparison on DSCNN KWS task. Parameters are reduced by $28\%$ for our method (frozen tail) and $30\%$ for magnitude pruning if maximum accuracy drop is set to $1.06\%$. Shallow networks limit pruning locations.}
    
    \label{figure:depthpruning_comp_kws}
\end{figure}
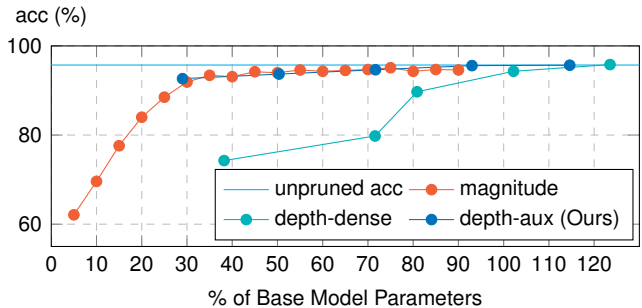

We compare our work to other pruning methods on VWW and KWS by visualizing the Pareto frontier in Figure \ref{figure:depthpruning_comp} and Figure \ref{figure:depthpruning_comp_kws} respectively. In both tasks, we are able to improve upon depth pruning with two dense layers using filter counts of $[64,32]$. On VWW, our method performs significantly better than global magnitude pruning with constant sparsity implemented using the TensorFlow Optimization Toolkit. Despite being at par with magnitude pruning on the KWS task, our method does not require sparse matrix support making it a better choice for on-device inference.

While our method has been shown to work on tinyML models, we believe it can be extended to architectures with a generally linear design such as plain ConvNets and ResNets. This is supported by work in intermediate layer analysis\cite{Alain2016} where a dense probe shows an increasing trend in separability in linear networks.

\subsection{Hardware Experiments}
We evaluate our work by deploying our pruned models on the Arducam Pico4ML running an ARM Cortex-M0 \cite{ARM2021} processor. Due to device limitations, the MobilenetV1 model was retrained using grayscale inputs. TFLite-Micro \cite{tensorflowLite2021} was used as the on-device inference framework and integer quantization was applied after pruning.

\begin{figure}
    \centering

    \begin{tikzpicture}[] 
    \begin{groupplot}[group style = {group size = 2 by 1, horizontal sep = 20pt}, width = 5.1cm, height = 4cm]

        \nextgroupplot[
            xlabel={Model Size (KB)},
            ymin=55, ymax=85,
            ytick scale label code/.code={},
            xtick scale label code/.code={},
            xtick={0,100000,200000,300000,400000},
            xticklabels={0,100,200,300,400},
            scatter,
            only marks,
            xmin=0,xmax=400000,
        	ylabel=acc (\%),
            colormap={cool}{rgb255(0cm)=(200,220,255); rgb255(1cm)=(0,128,255)},
            ymajorgrids=true,
            xmajorgrids=true,
            grid style=dashed,
            point meta=\thisrow{attachment_point},
            ticklabel style = {font=\scriptsize\sffamily\sansmath},
            label style = {font=\scriptsize\sffamily\sansmath},
            y label style = {at={(axis description cs:0,1.0)},rotate=-90,anchor=south},
            legend style = { column sep = 0pt, legend columns = -1, legend to name = grouplegend_hw,font=\scriptsize\sffamily\sansmath},
            legend cell align={left}
        ]
        \addplot[mark=triangle*, color={rgb:red,255;green,128;blue,0}, colormap={cool}{rgb255(0cm)=(255,220,200); rgb255(1cm)=(255,128,0)}]
            table[x=size,y=acc]{data/vww-dense_hardware.dat};
            \addlegendentry{depth-dense}
        \addplot[mark=*, color={rgb:red,0;green,128;blue,255}]
            table[x=size,y=acc]{data/vww_hardware.dat};
            \addlegendentry{depth-aux (Ours)}
            
        \draw[red, very thick, {Stealth}-{}] (axis cs:71000,81.11) -- (axis cs:100000,78);
        \node[align=left,font=\scriptsize\sffamily\sansmath\bfseries] at (axis cs:100000,78) [anchor=north west] {block6};
    
        \nextgroupplot[
            xlabel={Latency (ms)},
            ymin=55, ymax=85,
            ytick scale label code/.code={},
            xtick={100,300,500,700,900},
            scatter,
            only marks,
            xmin=100,xmax=950,
            colormap={cool}{rgb255(0cm)=(200,220,255); rgb255(1cm)=(0,128,255)},
        	ylabel=acc (\%),
            ymajorgrids=true,
            xmajorgrids=true,
            grid style=dashed,
            point meta=\thisrow{attachment_point},
            ticklabel style = {font=\scriptsize\sffamily\sansmath},
            label style = {font=\scriptsize\sffamily\sansmath},
            y label style = {at={(axis description cs:0,1.0)},rotate=-90,anchor=south}
        ]
        
        \addplot[mark=triangle*, color={rgb:red,255;green,128;blue,0}, colormap={cool}{rgb255(0cm)=(255,220,200); rgb255(1cm)=(255,128,0)}]
            table[x=latency,y=acc]{data/vww-dense_hardware.dat};    
        \addplot[mark=*]
            table[x=latency,y=acc]{data/vww_hardware.dat};

        \draw[red, very thick, {Stealth}-{}] (axis cs:551,81.11) -- (axis cs:625,78);
        \node[align=left,font=\scriptsize\sffamily\sansmath\bfseries] at (axis cs:625,78) [anchor=north west] {block6};
        
    \end{groupplot}
    \node at ($(group c2r1) + (-2.2cm,-2.5cm)$) {\ref{grouplegend_hw}}; 
    \end{tikzpicture} 

    \caption{VWW Depth Pruning on Cortex-M0. Darker marks denote deeper pruning points. Pruning at block6 of MobilenetV1-VWW reduces model size from 336KB to 71KB and latency from 904ms to 551ms. Accuracy also counterintuitively increases from $80\%$ to $81\%$.}
    \label{figure:depthpruning_hw}
\end{figure}
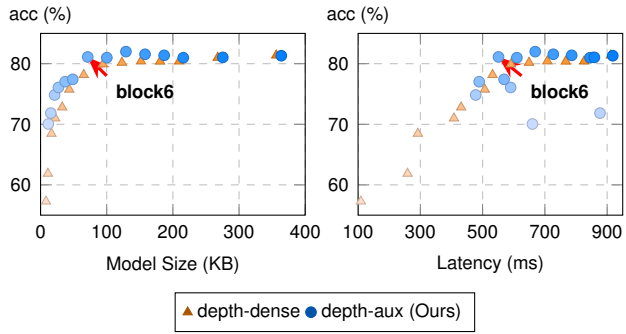

As shown in Figure \ref{figure:depthpruning_hw}, our approach using frozen tail training is able to produce a \texttt{tflite} model that is $4.7\times$ smaller in size which includes both parameters and architecture metadata. On-device inference latency is $1.6\times$ faster on pruned models with a slight accuracy boost of $1\%$. Pruning at block6 also produces a model that has $3\%$ higher accuracy compared to depth pruning using dense layers with a small size increase of $7$KB. Note that magnitude pruning cannot be deployed on-device due to lack of sparse matrix support.

\begin{figure}
    \centering

    \begin{tikzpicture}[] 
    \begin{groupplot}[group style = {group size = 2 by 1, horizontal sep = 20pt}, width = 5.1cm, height = 4cm]

        \nextgroupplot[
            xlabel={Model Size (KB)},
            ymin=40, ymax=100,
            ytick scale label code/.code={},
            xtick scale label code/.code={},
            ytick={40,50,60,70,80,90,100},
            xtick={10000,20000,30000,40000,50000,60000,70000},
            xticklabels={10,20,30,40,50,60,70},
            scatter,
            only marks,
            xmin=10000,xmax=70000,
        	ylabel=acc (\%),
            colormap={cool}{rgb255(0cm)=(200,220,255); rgb255(1cm)=(0,128,255)},
            ymajorgrids=true,
            xmajorgrids=true,
            grid style=dashed,
            point meta=\thisrow{attachment_point},
            ticklabel style = {font=\scriptsize\sffamily\sansmath},
            label style = {font=\scriptsize\sffamily\sansmath},
            y label style = {at={(axis description cs:0,1.0)},rotate=-90,anchor=south},
            legend style = { column sep = 0pt, legend columns = -1, legend to name = grouplegendKWS_hw,font=\scriptsize\sffamily\sansmath},
            legend cell align={left}
        ]
        \addplot[mark=triangle*, color={rgb:red,255;green,128;blue,0}, colormap={cool}{rgb255(0cm)=(255,220,200); rgb255(1cm)=(255,128,0)}]
            table[x=size,y=acc]{data/kws-dense_hardware.dat};
            \addlegendentry{depth-dense}
        \addplot[mark=*, color={rgb:red,0;green,128;blue,255}]
            table[x=size,y=acc]{data/kws_hardware.dat};
            \addlegendentry{depth-aux (Ours)}
            
        \draw[red, very thick, {Stealth}-{}] (axis cs:43840,90.77) -- (axis cs:50000,85);
        \node[align=left,font=\scriptsize\sffamily\sansmath\bfseries] at (axis cs:50000,85) [anchor=north west] {block2};
    
        \nextgroupplot[
            xlabel={Latency (ms)},
            ymin=40, ymax=100,
            ytick scale label code/.code={},
            xtick={0,100,200,300,400},
            ytick={40,50,60,70,80,90,100},
            scatter,
            only marks,
            xmin=50,xmax=450,
            colormap={cool}{rgb255(0cm)=(200,220,255); rgb255(1cm)=(0,128,255)},
        	ylabel=acc (\%),
            ymajorgrids=true,
            xmajorgrids=true,
            grid style=dashed,
            point meta=\thisrow{attachment_point},
            ticklabel style = {font=\scriptsize\sffamily\sansmath},
            label style = {font=\scriptsize\sffamily\sansmath},
            y label style = {at={(axis description cs:0,1.0)},rotate=-90,anchor=south}
        ]
        
        \addplot[mark=triangle*, color={rgb:red,255;green,128;blue,0}, colormap={cool}{rgb255(0cm)=(255,220,200); rgb255(1cm)=(255,128,0)}]
            table[x=latency,y=acc]{data/kws-dense_hardware.dat};    
        \addplot[mark=*]
            table[x=latency,y=acc]{data/kws_hardware.dat};

        \draw[red, very thick, {Stealth}-{}] (axis cs:275,90.77) -- (axis cs:310,85);
        \node[align=left,font=\scriptsize\sffamily\sansmath\bfseries] at (axis cs:310,85) [anchor=north west] {block2};
        
    \end{groupplot}
    \node at ($(group c2r1) + (-2.2cm,-2.5cm)$) {\ref{grouplegendKWS_hw}}; 
    \end{tikzpicture} 

    \caption[KWS Depth Pruning on Cortex-M0]{KWS Depth Pruning Results on Cortex-M0. Darker marks denote deeper pruning points. Pruning at block2 of DSCNN reduces model size from 54KB to 43KB and latency from 340ms to 275ms at a cost of $2.21\%$ accuracy.}
    \label{figure:depthpruningKWS_hw}
\end{figure}
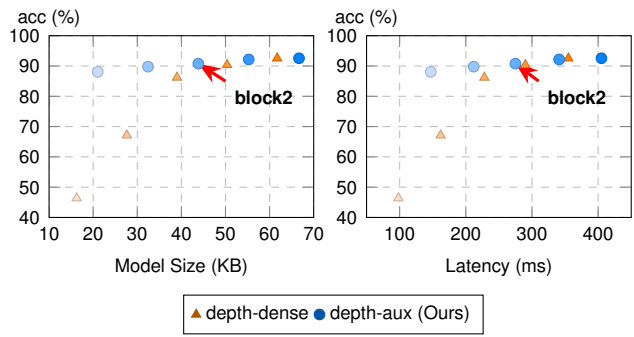

We also validate our results on the KWS task with the DSCNN model \cite{Zhang2017} on the Cortex-M0. We first retrain the base model using Log Mel-filterbank Energies (LFBE) instead of Mel-frequency cepstral coefficients (MFCC) as the input preprocessing in order to use optimized on-device libraries. As shown in Figure \ref{figure:depthpruningKWS_hw}, our results improve model size by by $1.2\times$ and latency by $1.2\times$ at the cost of $2.21\%$ accuracy when pruned at block2.

%% file: sections/5-conclusion.tex
In this paper we introduce our depth pruning method which uses an auxiliary network as a new head of the pruned model. This technique is easily interpretable and requires no special hardware support during inference. Results show that we are able to significantly reduce model size with minimal accuracy loss in tinyML tasks. On-device experiments validate our work with significant speedup and reduction to memory footprint which translates to energy savings in IoT devices.

These results open up new avenues for optimization to satisfy extreme hardware constraints. In future work, we aim to apply this technique to dynamic architectures opening up new opportunities for efficient on-device inference.